\documentclass[journal=acscii,manuscript=article]{achemso}
\setkeys{acs}{usetitle=true}

\usepackage[version=3]{mhchem} % Formula subscripts using \ce{}
\usepackage[T1]{fontenc}       % Use modern font encodings
\usepackage{epstopdf}
\usepackage[utf8]{inputenc} % allow utf-8 input
\usepackage{hyperref}       % hyperlinks
\usepackage{url}            % simple URL typesetting
\usepackage{booktabs}       % professional-quality tables
\usepackage{amsfonts}       % blackboard math symbols
\usepackage{nicefrac}       % compact symbols for 1/2, etc.
\usepackage{microtype}      % microtypography
\usepackage{graphicx}
\usepackage{subcaption}
\usepackage[subrefformat=parens,labelformat=parens]{subcaption}
\usepackage{algorithm}
\usepackage[noend]{algpseudocode}
\usepackage{color,soul}
\usepackage{amsmath,amsthm,amssymb,amsfonts}
\usepackage{mathtools}
\mathtoolsset{showonlyrefs}
\usepackage{placeins}
\usepackage[T1]{fontenc}
\usepackage{lmodern}

\newcommand{\sidecaption}[1]% #1 = label name
{\raisebox{\abovecaptionskip}{\begin{subfigure}[t]{1.6em}
  \caption[singlelinecheck=off]{}% do not center
  \label{#1}
\end{subfigure}}\ignorespaces}

\algrenewcommand{\algorithmicrequire}{\textbf{Input:}}
\algrenewcommand{\algorithmicensure}{\textbf{Output:}}
\algrenewcommand\Return{\State\algorithmicreturn{}~}
\algrenewcommand\algorithmicindent{1.0em}%

\author{Rafael G\'omez-Bombarelli}
\affiliation{Kyulux North America Inc.}
\altaffiliation{Equal contribution}

\author{Jennifer N. Wei}
\affiliation{Department of Chemistry and Chemical Biology, Harvard University, Cambridge MA 02138, USA}
\altaffiliation{Equal contribution}

\author{David Duvenaud}
\affiliation{Department of Computer Science, University of Toronto}
\altaffiliation{Equal contribution}

\author{Jos\'e Miguel Hern\'andez-Lobato}
\affiliation{Department of Engineering, University of Cambridge Trumpington Street, Cambridge CB2 1PZ, UK}
\altaffiliation{Equal contribution}

\author{Benjam\'in S\'anchez-Lengeling}
\author{Dennis Sheberla}
\affiliation{Department of Chemistry and Chemical Biology, Harvard University, Cambridge MA 02138, USA}

\author{Jorge Aguilera-Iparraguirre}
\affiliation{Kyulux North America Inc.}

\author{Timothy D. Hirzel}
\affiliation{Kyulux North America Inc.}

\author{Ryan P. Adams}
\affiliation{Google Brain and Princeton University}

\author{Al\'an Aspuru-Guzik}
\email{aspuru@chemistry.harvard.edu}
\affiliation{Department of Chemistry and Chemical Biology, Harvard University, Cambridge MA 02138, USA}
\alsoaffiliation{Canadian Institute for Advanced Research (CIFAR), Biologically-Inspired Solar Energy Program.}

\title{Automatic Chemical Design Using a Data-Driven Continuous Representation of Molecules} 

\begin{document}

\begin{abstract}
We report a method to convert discrete representations of molecules to and from a multidimensional continuous representation.
This model allows us to generate new molecules for efficient exploration and optimization through open-ended spaces of chemical compounds.

A deep neural network was trained on hundreds of thousands of existing chemical structures to construct three coupled functions: an encoder, a decoder and a predictor. The encoder converts the discrete representation of a molecule into a real-valued continuous vector, and the decoder converts these continuous vectors back to discrete molecular representations. The predictor estimates chemical properties from the latent continuous vector representation of the molecule.

Continuous representations allow us to automatically generate novel chemical structures by performing simple operations in the latent space, such as decoding random vectors, perturbing known chemical structures, or interpolating between molecules.

Continuous representations also allow the use of powerful gradient-based optimization to efficiently guide the search for optimized functional compounds.
We demonstrate our method in the domain of drug-like molecules and also in the set of molecules with fewer that nine heavy atoms.
\end{abstract}

\section{Introduction}

The goal of drug and material design is to identify novel molecules that have certain desirable properties.  We view this as an optimization problem, in which we are searching for the molecules that maximize our quantitative desiderata.
However, optimization in molecular space is extremely challenging, because the search space is large, discrete, and unstructured.
Making and testing new compounds is costly and time consuming, and the number of potential candidates is overwhelming. Only about $10^8$ substances have ever been synthesized, \cite{Kim2016} whereas the range of potential drug-like molecules is estimated to be between $10^{23}$ and $10^{60}$.\cite{polischuk2013}

Virtual screening can be used to speed up this search.\cite{shoichet_2004_virtual,scior2012,cheng2012,Pizza2015} Virtual libraries containing thousands to hundreds of millions of candidates can be assayed with first-principles simulations or statistical predictions based on learned proxy models, and only the most promising leads are selected and tested experimentally.

However, even when accurate simulations are available,\cite{schneider_2010_virtual} computational molecular design is limited by the search strategy used to explore chemical space.
Current methods either exhaustively search through a fixed library,\cite{hachmann2011harvard,bombarelli2016} or use discrete local search methods such as genetic algorithms\cite{Virshup_2013, Rupakheti_2015, Reymond_2015, Reymond_2010,kanal_2013_efficient,oboyle_2011_computational} or similar discrete interpolation techniques.\cite{van_Deursen_2007,wang2006designing,balamurugan2008exploring}
Although these techniques have led to useful new molecules, these approaches still face large challenges. Fixed libraries are monolithic, costly to fully explore, and require hand-crafted rules to avoid impractical chemistries. The genetic generation of compounds requires the manual specification of heuristics for mutation and crossover rules. Discrete optimization methods have difficulty effectively searching large areas of chemical space because since it is not possible guide the search with gradients. 

A molecular representation method that is continuous, data-driven, and can easily be converted into a machine-readable molecule has several advantages. 
First, hand-specified mutation rules are unnecessary, as new compounds can be generated automatically by modifying the vector representation and then decoding.
Second, if we develop a differentiable model that maps from molecular representations to desirable properties, we can enable the use of gradient-based optimization to make larger jumps in chemical space. 
Gradient-based optimization can be combined with Bayesian optimization methods to select compounds that are likely to be informative about the global optimum.
Third, a data-driven representation can leverage large sets of unlabeled chemical compounds to automatically build an even larger implicit library, and then use the smaller set of labeled examples to build a regression model from the continuous representation to the desired properties.
This lets us take advantage of large chemical databases containing millions of molecules, even when many properties are unknown for most compounds.

Recent advances in machine learning have resulted in powerful probabilistic generative models that, after being trained on real examples, are able to produce realistic synthetic samples. % that are perceived as valid by humans.
Such models usually also produce low-dimensional continuous representations of the data being modeled, allowing interpolation or analogical reasoning for 
natural images\cite{radford2015unsupervised}, text\cite{bowman2015generating}, speech, and music\cite{vandenoord_2016}.
We apply such generative models to chemical design, using a pair of deep networks trained as an autoencoder to convert molecules represented as SMILES strings into a continuous vector representation. In principle, this method of converting from a molecular representation to a continuous vector representation could be applied to any molecular representation, including 
chemical fingerprints,\cite{ECFP2010} 
convolutional neural networks on graphs\cite{duvenaud2015convolutional},
 similar graph-convolutions\cite{kearnes_2016_molecular}, and 
 Coulomb matrices \cite{Rupp_2012}. We chose to use SMILES representation because it can be readily converted into a molecule.

Using this new continuous vector-valued representation, we experiment with the use of continuous optimization to produce novel compounds. We trained the autoencoder jointly on a property prediction task; we added a multilayer perceptron that predicts property values from the continuous representation generated by the encoder and included the regression error in our loss function. We examined the effects this joint training had on the latent space.

\begin{figure}[h]
\begin{center}
\includegraphics[width=0.90\textwidth]{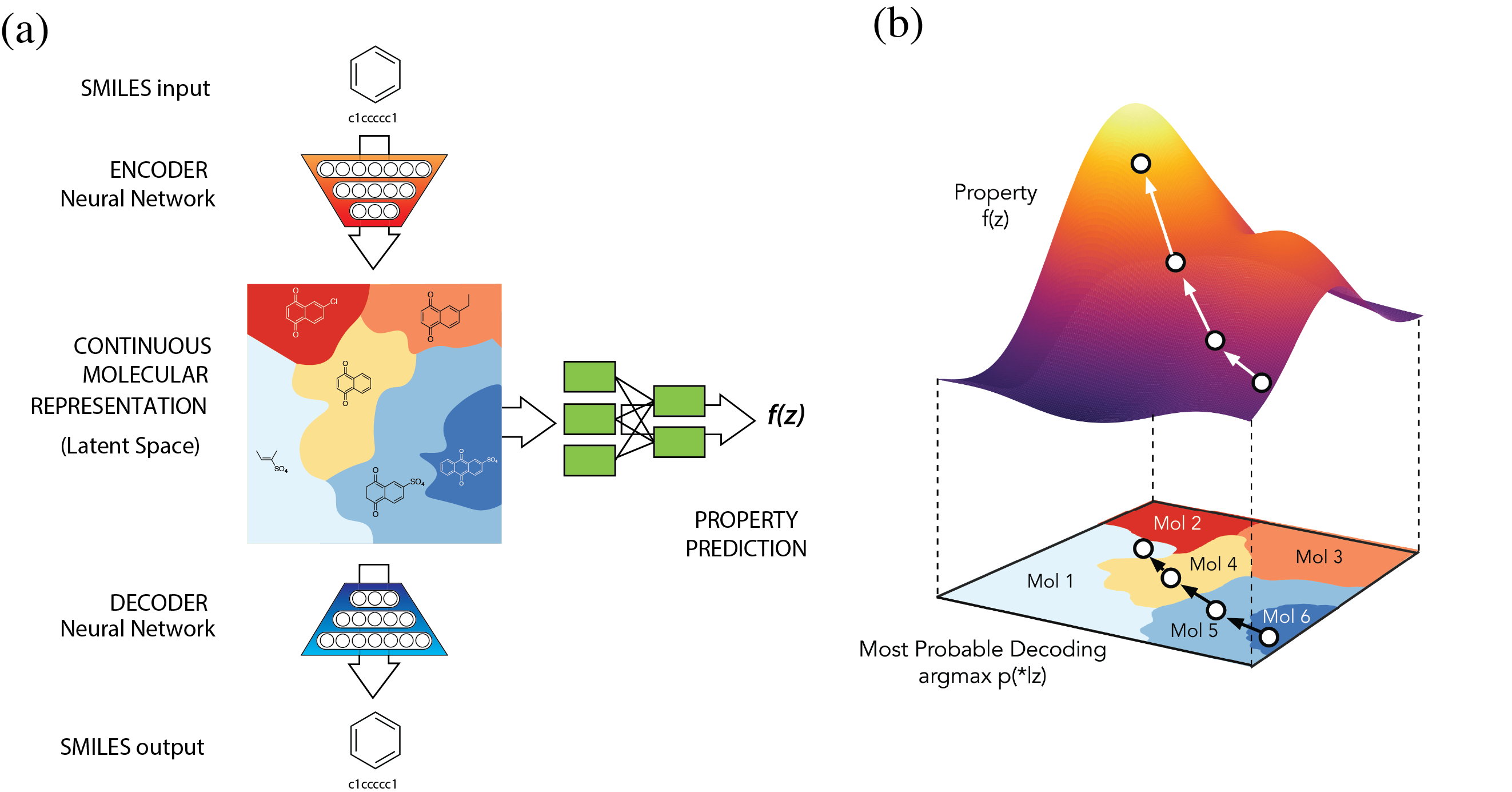} 
\caption{\small	\textbf{(a)}. A diagram of the proposed autoencoder for molecular design, including the joint property prediction model.
Starting from a discrete molecular representation, such as a SMILES string, the encoder network converts each molecule into a vector in the latent space, which is effectively a continuous molecular representation. Given a point in the latent space, the decoder network produces a corresponding SMILES string. Another network estimates the value of target properties associated with each molecule.
\textbf{(b)} Gradient-based optimization in continuous latent space.
After training a surrogate model $f(z)$ to predict the properties of molecules based on their latent representation $z$, we can optimize $f(z)$ with respect to $z$ to find new latent representations expected to have high values of desired properties.
These new latent representations can then be decoded into SMILES strings, at which point their properties can be tested empirically.}
\label{fig:ae_opt_diagrams}
\end{center}
\end{figure}

\subsection{Representation and Autoencoder Framework}
\subsection{Initial representation of molecules}%the types of encoded and decoded molecules}
Before building an encoder that produces a continuous latent representation, we must choose which discrete molecular representation to use for the input and output.
To leverage the power of recent advances in sequence-to-sequence autoencoders for modeling text\cite{bowman2015generating}, we used the SMILES\cite{Weininger_1988} representation, a commonly-used text encoding for organic molecules.
We also tested InChI\cite{heller_2013_inchi} as an alternative string representation, but found it to perform substantially worse than SMILES, presumably due to a more complex syntax that includes counting and arithmetic.

\subsection{Training an autoencoder}
Starting from a large library of string-based representations of molecules, we trained a pair of deep neural networks:
an encoder network to convert each string into a fixed-dimensional vector,
and a decoder network to convert vectors back into strings (Figure \ref{fig:ae_opt_diagrams}a).
Such encoder-decoder pairs are known as \textit{autoencoders}.
The autoencoder is trained to minimize error in reproducing the original string, \textit{i.e.}, it attempts to learn the identity function.
Key to the design of the autoencoder is mapping through an \emph{information bottleneck}.
This bottleneck --- here the fixed-length continuous vector --- induces the network to learn a compressed representation that captures the most statistically salient information in the data.
We call the vector-encoded molecule the \emph{latent representation} of the molecule.

The character-by-character nature of the SMILES representation and the fragility of its internal syntax (opening and closing cycles and branches, allowed valences, etc.) can result in the output of invalid molecules from the decoder. 
Multiple factors contribute to the proportion of valid SMILES output from the decoder, including atom count and training set density. In generating new SMILES strings, the percentage of valid SMILES output ranged from 70\% to less than 1\%. We employed the open source cheminformatics suite RDKit\cite{rdkit} and Marvin to validate the chemical structures of output molecules and discard invalid ones. While it would be more efficient to have the autoencoder generate only valid strings, this post-processing step is lightweight and allows for greater flexibility in the autoencoder to learn the architecture of the SMILES.

To enable molecular design, the chemical structures encoded in the continuous representation of the autoencoder need to be correlated with the target properties that we are seeking to optimize. Therefore, we added a model to the autoencoder that predicts the properties from the latent space representation. This autoencoder was then trained jointly on the reconstruction task and a property prediction task; an additional multi-layer perceptron (MLP) was used to predict the property from the latent vector of the encoded molecule. To propose promising new candidate molecules, we can start from the latent vector of an encoded molecule and then move in the direction most likely to improve the desired attribute. The resulting new candidate vectors can then be decoded into corresponding molecules. (Figure \ref{fig:ae_opt_diagrams}b)

\subsection{Using variational autoencoders to produce a latent representation.}
For unconstrained optimization in the latent space to work, points in the latent space must decode into valid SMILES strings that capture the chemical nature of the training data.
However, the original training objective of the autoencoder does not enforce this constraint, as we chose to handle invalid SMILES in a post processing step. As a result, the latent space learned by the autoencoder may contain large ``dead areas'', which decode to invalid SMILES strings.

To ensure that points in the latent space correspond to valid realistic molecules, we modified our autoencoder and its objective into a \emph{variational} autoencoder (VAE)~\cite{kingma2013auto}.
VAEs were developed as a principled approximate-inference method for latent-variable models, in which each datum has a corresponding, but unknown, latent representation.
VAEs generalize autoencoders, adding stochasticity to the encoder which combined with a penalty term encourages all areas of the latent space to correspond to a valid decoding.
The intuition is that adding noise to the encoded molecules forces the decoder to learn how to decode a wider variety of latent points and find more robust representations.
In addition, since two different molecules can have their encodings stochastically brought close in the latent space, but still need to decode to different molecular graphs, this constraint encourages the encodings to spread out over the entire latent space to avoid overlap.
Variational autoencoders with recurrent neural network encoding/decoding were proposed by Bowman \textit{et al.} in the context of written English sentences and we followed their approach closely.\cite{bowman2015generating}

Two autoencoder system were trained; one with 108,000 molecules from the QM9 dataset of molecules with fewer than 9 heavy atoms \cite{ramakrishnan2014quantum} and another with 250,000 drug-like commercially available molecules extracted at random from the ZINC database.\cite{irwin_2012_zinc}

We performed random optimization over hyperparameters specifying the deep autoencoder architecture and training, such as the choice between a recurrent or convolutional encoder, the number of hidden layers, layer sizes, regularization and learning rates. 
The latent space representations for the QM9 and ZINC datasets had 156 dimensions and 196 dimensions respectively.

%%%%%%%%%%%%%%

\section{Results and discussion}
\begin{figure}[ht]
	\centering
	\sidecaption{subfig:a}
	\raisebox{-\height}{\includegraphics[width=0.45\linewidth]{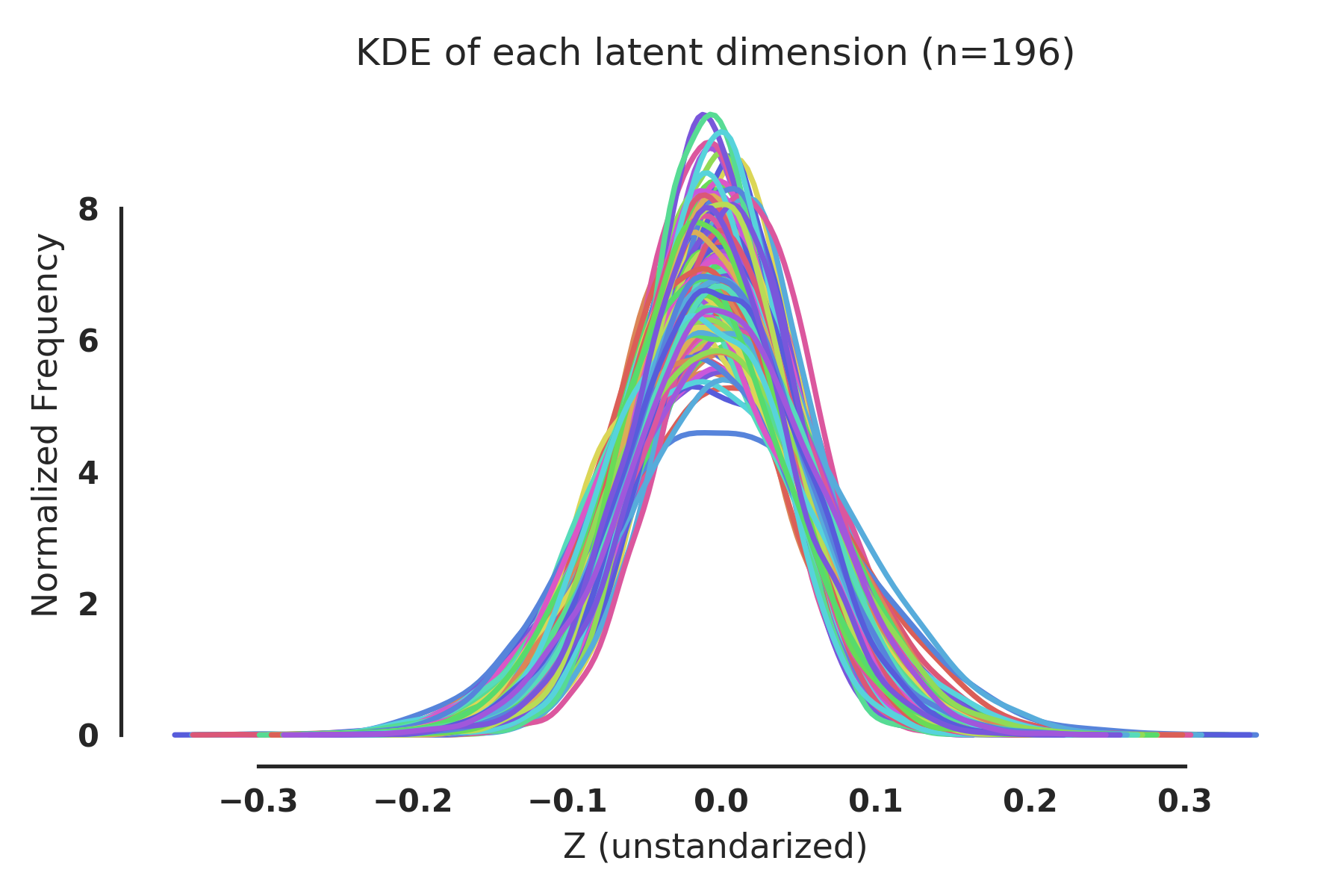}}
	\hfill
	\sidecaption{subfig:b}
	\raisebox{-\height}{\includegraphics[width=0.45\linewidth]{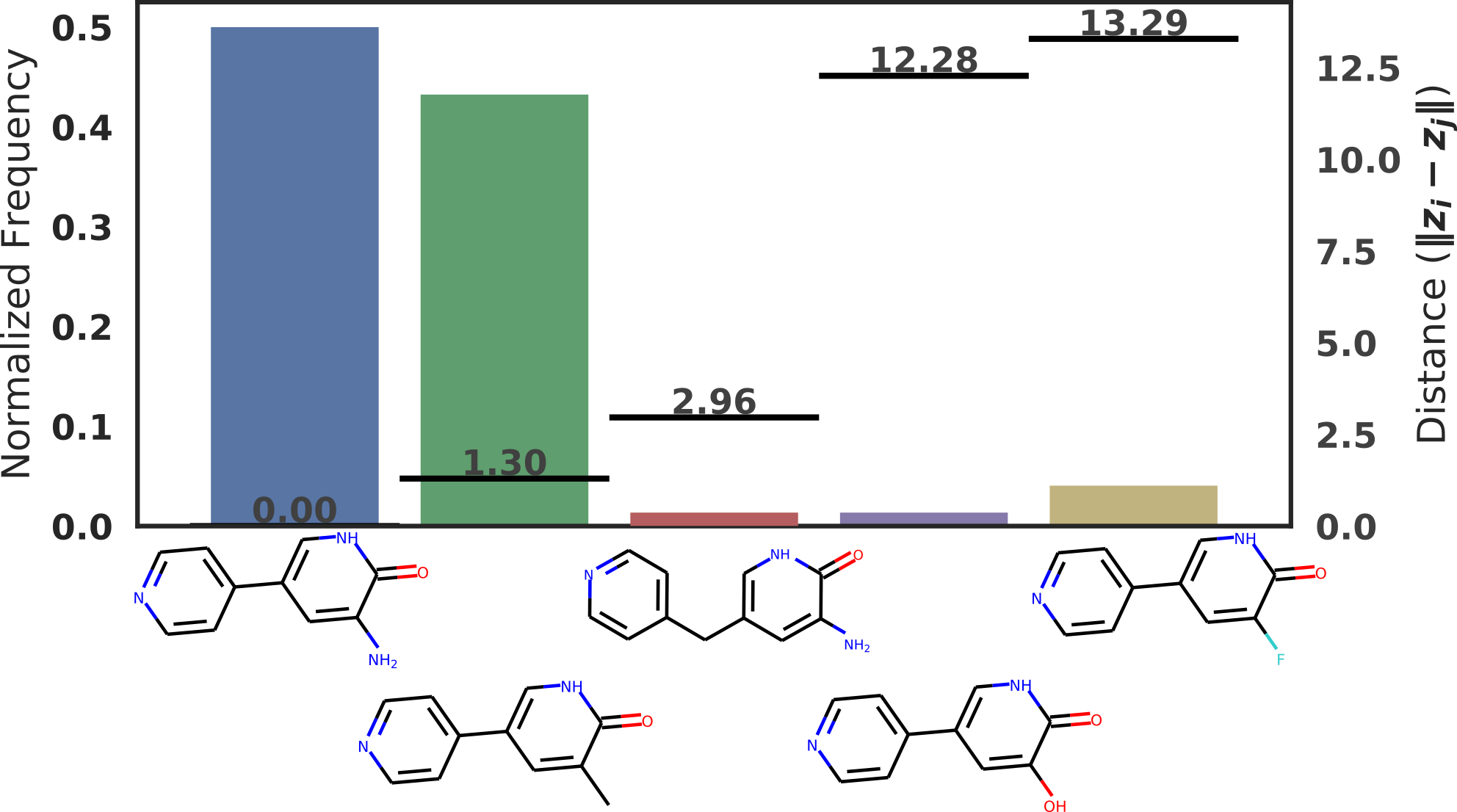}}  
	
	\vspace{1.5cm}
	
	\sidecaption{subfig:c}
	\raisebox{-\height}{\includegraphics[width=\linewidth]{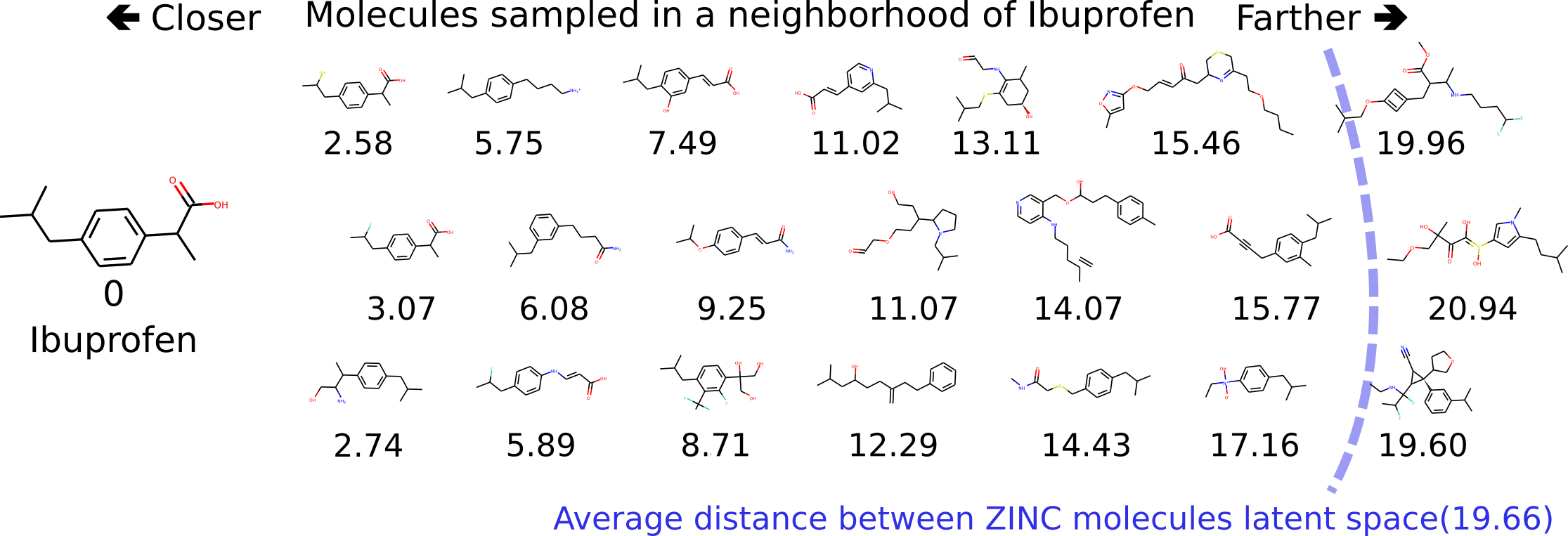}} 
	
	\vspace{1.5cm}
	
	\sidecaption{subfig:d}	
	\raisebox{-\height}{\includegraphics[width=\linewidth]{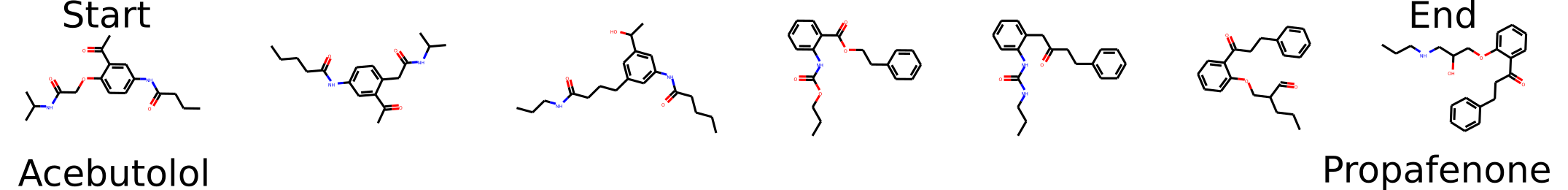}} 	

    \caption{Representations of the sampling results from the variational autoencoder. (a) Kernel Density Estimation (KDE) of each latent dimension of the autoencoder; (b) Histogram of sampled molecules for a single point in the latent space, the distances of the molecules from the original query are shown by the lines corresponding to the right axis; (c) Molecules sampled near the location of ibuprofen in latent space. The values below the molecules are the distance in latent space from the decoded molecule to ibuprofen; (d) \textit{slerp} interpolation between two molecules in latent space using 6 steps of equal distance.} 
    \label{fig:interpolation_figures}
\end{figure}

\paragraph{Representation of molecules in latent space}
Firstly, we analyze the fidelity of the autoencoder and the ability of the latent space to capture structural molecular features.
Figure \ref{fig:interpolation_figures}a) shows a kernel density estimate of each dimension when encoding a set of 5000 randomly selected ZINC molecules from outside the training set. 
 Whereas each individual dimension shows a slightly different mean and standard deviation, all follow normal distribution as enforced by the variational regularizer. 

The variational autoencoder is a doubly-probabilistic model. In addition to the added noise to the encoder, which can be turned off by simply sampling the mean of the encoding distribution, the decoder also samples a string from of the probability distribution over characters in each position generated by the final layer. This implies that decoding a single point in the latent space back to a string representation is stochastic. Figure \ref{fig:interpolation_figures}b) shows the probability of decoding the latent representation of a sample FDA-approved drug molecule into several different molecules. For most latent points, a prominent molecule is decoded and many other slight variations appear with lower frequencies. When these resulting SMILES are re-encoded into the latent space, the most frequent decoding also tends to be the one with the lowest Euclidean distance to the original point, indicating the latent space is indeed capturing features relevant to molecules.

Figure \ref{fig:interpolation_figures}c) shows some molecules in the latent space that are close to ibuprofen. These structures become less similar with increasing distance in the latent space. When the distance approaches the average distance of molecules in the training set, the changes are more pronounced, eventually resembling random molecules likely to be sampled from the training set. Figure 5d) shows the distribution of distances in latent space between 50,000 random points from our ZINC training set.

A continuous latent space allows interpolation of molecules by following the shortest Euclidean path between their latent representations. When exploring high dimensional spaces, it is important to note that Euclidean distance might not map directly to notions of similarity of molecules \cite{Aggarwal2001}. In high dimensional spaces, most of the mass of independent normally distributed random variables is not near the mean, but in an increasingly distant annulus around it \cite{Domingos2012}. Interpolating linearly between two points might pass by an area of low probability, to keep the sampling on the areas of high probability we utilize spherical interpolation\cite{White2016} (\emph{slerp}). With \emph{slerp}, the path between two points is a circular arc lying on the on the surface of a N-dimensional sphere. Figure \ref{fig:interpolation_figures}d) shows the spherical interpolation between two random drug molecules, showing smooth transitions in between. Figure 7 shows the difference between linear and spherical interpolation.

\begin{table}[h]
\small
\centering
%\begin{tabular}{lclclclclccc}
\begin{tabular}{lp{1cm}lp{1.3cm}lp{1cm}lp{1cm}lp{1cm}lp{1cm}lp{1cm}l}
\hline
  Source$^{a}$ & Dataset$^{b}$ &  Samples$^{c}$ &   logP$^{d}$ &      SAS$^{e}$  &   QED$^{f}$ & \% in ZINC$^{g}$ & \% in emol$^{h}$ \\
\hline
   Data &    ZINC &     249k &  2.46 (1.43) &  3.05 (0.83)  &  0.73 (0.14) & 100 & 12.9 \\
  GA &    ZINC &     5303 &  2.84 (1.86) &  3.80 (1.01) &  -0.82 (0.71) &        6.5 &  4.8\\
    VAE &    ZINC &     8728 &  2.67 (1.46) &  3.18 (0.86) &  -0.96 (0.75) &          4.5 & 7.0\\
\hline
Data &     QM9 &     134k &  0.31 (1.00) &  4.24 (0.91) &  0.99 (1.20)  &    0.0 & 8.6 \\
GA &     QM9 &     5470 &  0.96 (1.53) &  4.47 (1.01) &  0.68 (0.97)  &        0.018    & 3.8 \\
 VAE &     QM9 &     2839 &  0.30 (0.97) &  4.34 (0.98) &  0.47 (0.08) &        0.0  &  8.9\\
 \hline
 \end{tabular}
 \caption{\small	Comparison of molecule generation results using genetic algorithm (GA) and variational autoencoder (VAE) without joint property prediction. 
a) Describes the source of the molecules, Data refers to the entire dataset; b) The dataset used, either ZINC or QM9, 
c) Number of samples used fro comparison; d) water-octanal partition coefficient (\texttt{logP})\cite{wildman_1999_prediction}; e) synthetic accessibility score (\texttt{SAS})\cite{Ertl2009estimation}; f) Qualitative Estimate of Drug-likeness (\texttt{QED})\cite{bickerton2012quantifying}; g) percentage of generated molecules found in ZINC; h) percentage of generated molecules founds in E-molecules (emol) database\cite{emolecules}}
\label{tab:zinc_gen_results}
 \end{table}

Table \ref{tab:zinc_gen_results} compares the distribution of chemical properties in the training sets with a) random molecules generated with a list of hand-designed rules\cite{Virshup_2013, Rupakheti_2015, Reymond_2015, Reymond_2010,kanal_2013_efficient,oboyle_2011_computational} and b) with molecules decoded from sampling random points in the latent space of an VAE trained only for the reconstruction task. We compare the water-octanol partition coefficient (logP), the synthetic accessibility score (SAS), the natural-product score (NP) and drug-likeness (QED). Despite the fact that the VAE is trained purely on the SMILES strings independently of chemical properties, it is able to generate realistic-looking molecules whose features follow the intrinsic distribution of the training data. The two rightmost columns in Table \ref{tab:zinc_gen_results} report the fraction of molecules that belong to the the 17 million drug-like compounds from which the training set was selected and how often they can be found in a library of existing organic compounds. In the case of drug-like molecules, the VAE generates molecules that follow the property distribution of the training data, but are new, since the combinatorial space is extremely large the training set is an arbitrary sub-sample. The hand-selected mutations are less able to generate new compounds while at the same time biasing the properties of the set to higher chemical complexity and decreased drug-likeness. In the case of the QM9 dataset, since the combinatorial space is smaller, the training set has more coverage and the VAE generates essentially the same population statistics as the training data.
 
\paragraph{Property prediction of molecules}
The interest in discovering new molecules and chemicals is most often in relation to maximizing some desirable property. For this reason, we extended the the purely generative model to also predict property values from the latent representation. We trained a multi-layer perceptron jointly with the autoencoder to predict properties from the latent representation of each molecule.

\begin{table}[ht]
\centering
\begin{tabular}{|c|c|c|c|c|c|c|c|}
\hline
Database/Property & Mean$^{a}$   & ECFP$^{b}$    & CM$^{b}$ & GC$^{b}$ &  1-hot SMILES$^{c}$ & Encoder$^{d}$ & VAE$^{e}$   \\ %check if needs to be updated
\hline
ZINC250k/logP & 1.14 & 0.38 & - & 0.05 & 0.16 & 0.13 & 0.15 \\ %& 0.18  \\  
ZINC250k/QED & 0.112 & 0.045 & - & 0.017 & 0.041 & 0.037 & 0.054 \\ %0.030  \\
QM9/HOMO, eV & 0.44 & 0.20 & 0.16 & 0.12 & 0.12 & 0.13 & 0.16  \\ %0.13  \\
QM9/LUMO, eV & 1.05 & 0.20 & 0.16 & 0.15 & 0.11 & 0.14 & 0.16  \\
QM9/Gap, eV & 1.07 & 0.30 & 0.24 & 0.18 & 0.16 & 0.18 & 0.21  \\ 
\hline 
\end{tabular} 
\caption{\small	MAE prediction error for  properties using various methods on the ZINC and QM9 datasets. 
a) Baseline, mean prediction; b) As implemented in Deepchem benchmark (MoleculeNet)\cite{wu2017moleculenet}, ECFP-circular fingerprints, CM-coulomb matrix, GC-graph convolutions; c) 1-hot-encoding of SMILES used as input to property predictor; d) The network trained without decoder loss; e) full variational autoencoder network trained for individual properties.}
\label{tab:props_mae}
\end{table}

\begin{figure}[ht]
    \centering
     \includegraphics[width=\textwidth]{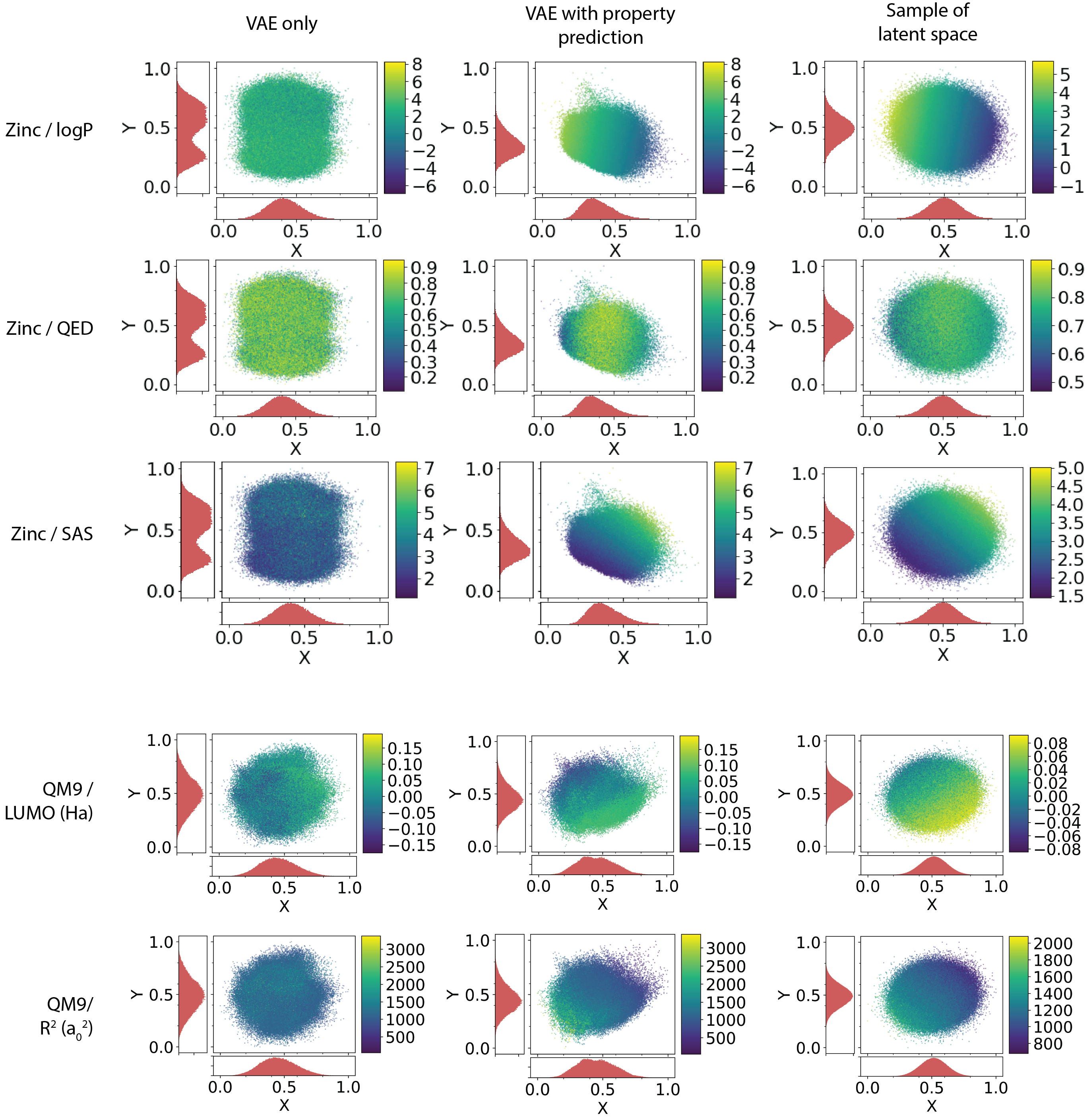}
    \caption{Two-dimensional PCA analysis of latent space for variational autoencoder. The two axis are the principle components selected from the PCA analysis, the color bar shows the value of the selected property. The first column shows the representation of all molecules from the listed dataset using autoencoders trained without joint property prediction. The second column shows the representation of molecules using an autoencoder trained with joint property prediction. The third column shows a representation of random points in the latent space of the autoencoder trained with joint property prediction; the property values predicted for these points are predicted using the property predictor network. The first three rows show the results of training on molecules from the ZINC dataset for the \texttt{logP}, \texttt{QED}, and \texttt{SAS} properties; the last two rows show the results of training on the QM9 dataset for the LUMO energy and the electronic spatial extent (R$^2$). } \label{fig:ls_figures}
    \vspace{1cm}    
\end{figure}

With joint training for property prediction, the distribution of molecules in the latent space is organized by property values. Figure \ref{fig:ls_figures} shows the mapping of true property values to the latent space representation of molecules, compressed into two dimensions using PCA. The latent space generated by autoencoders jointly trained with the property prediction task shows in the distribution of molecules a gradient by property values; molecules with high values are located in one region, and molecules with  low values in another. Autoencoders that were trained without the property prediction task do not show a discernible pattern with respect to property values in the resulting latent representation distribution. 

While the primary purpose of adding property prediction was to organize the latent space, it is interesting to observe how the property predictor model compares with other standard models for property prediction. Table \ref{tab:props_mae} compares the performance of commonly used molecular embeddings and models to the VAE. Our VAE model shows that property prediction performance for electronic properties (\textit{i.e.}, orbital energies) are similar to graph convolutions for some properties; prediction accuracy could be improved with further hyperparameter optimization.

\paragraph{Optimization of molecules via properties}

We next optimized molecules in the latent space from the autoencoder which was jointly trained for property prediction. We used a Gaussian process model\cite{rasmussen2006gaussian} to predict target properties for molecules given the latent space representation of the molecules as an input. The 2,000 molecules used for training the Gaussian process were selected to be maximally diverse. Using this model, we optimized in the latent space to find a molecule that maximized our objective. As a baseline, we compared our optimization results against molecules found using a random Gaussian search and molecules optimized via a genetic algorithm.

The objective we chose to optimize was $5 \times$ \texttt{QED} $-$ \texttt{SAS}, where \texttt{QED} is the Quantitative Estimation of Drug-likeness (QED)\cite{bickerton2012quantifying}, and \texttt{SAS} is the Synthetic Accessibility score\cite{Ertl2009estimation}. This objective represents a rough estimate of finding the most drug-like molecule that is also easy to synthesize. To provide the greatest challenge for our optimizer, we started with molecules from the ZINC dataset that were in the bottom 10\% percentile of our objective.

\begin{figure}[ht]
	\centering
	\sidecaption{subfig:a}
	\raisebox{-\height}{\includegraphics[width=0.45\linewidth]{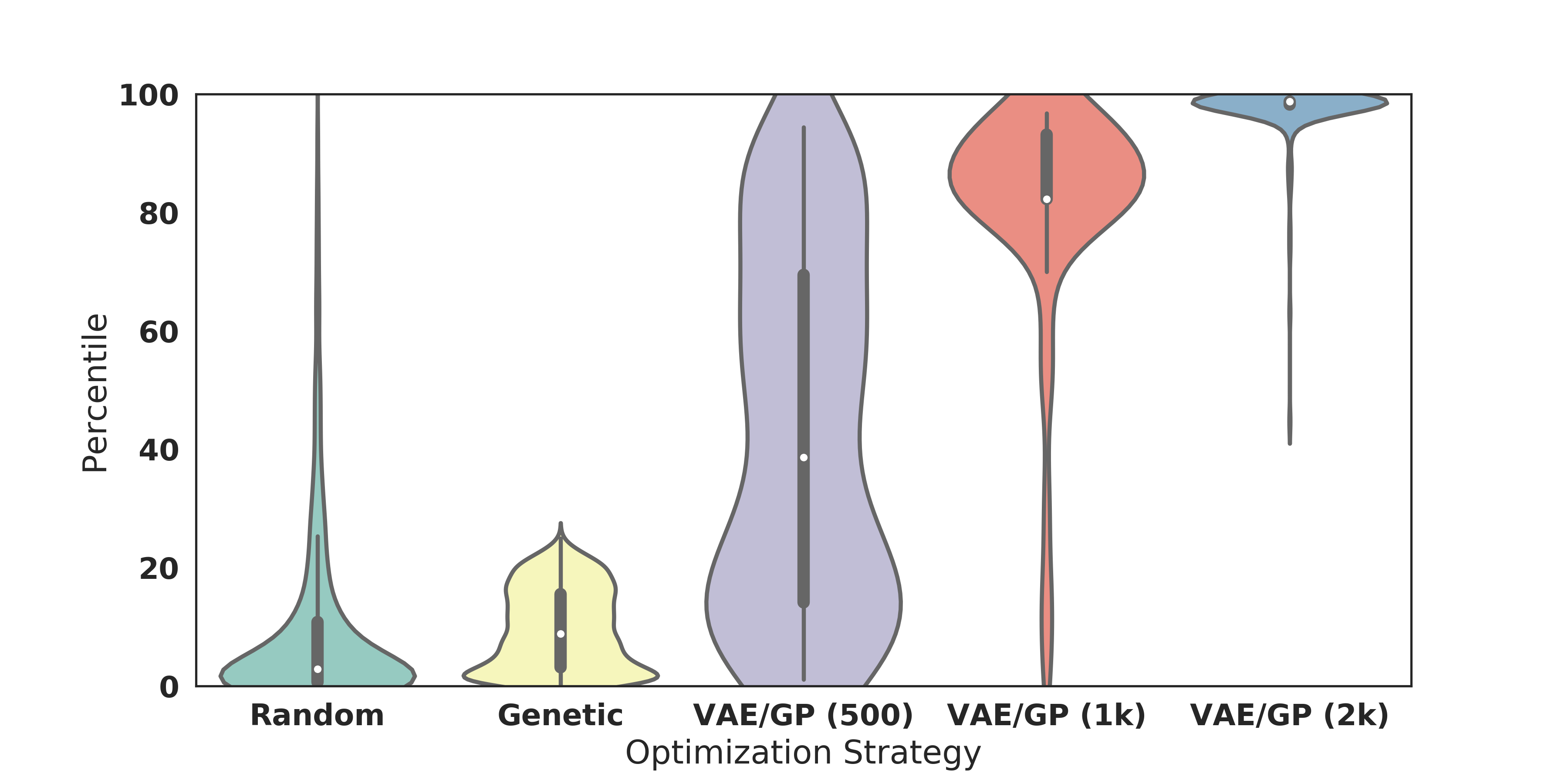}}
	\sidecaption{subfig:b}
	\raisebox{-\height}{\includegraphics[width=0.45\linewidth]{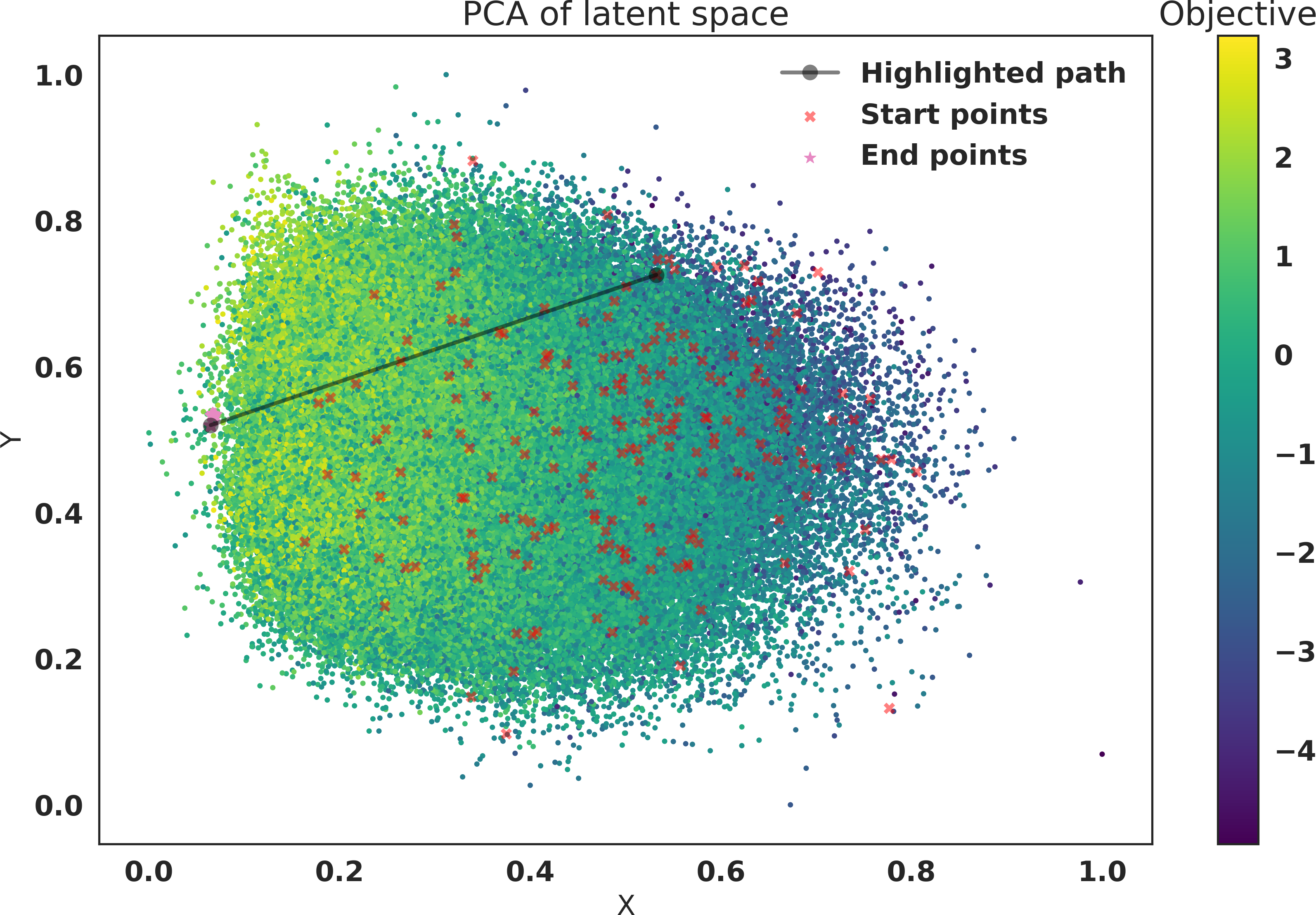}} 

	\sidecaption{subfig:c}
	\raisebox{-\height}{\includegraphics[width=\linewidth]{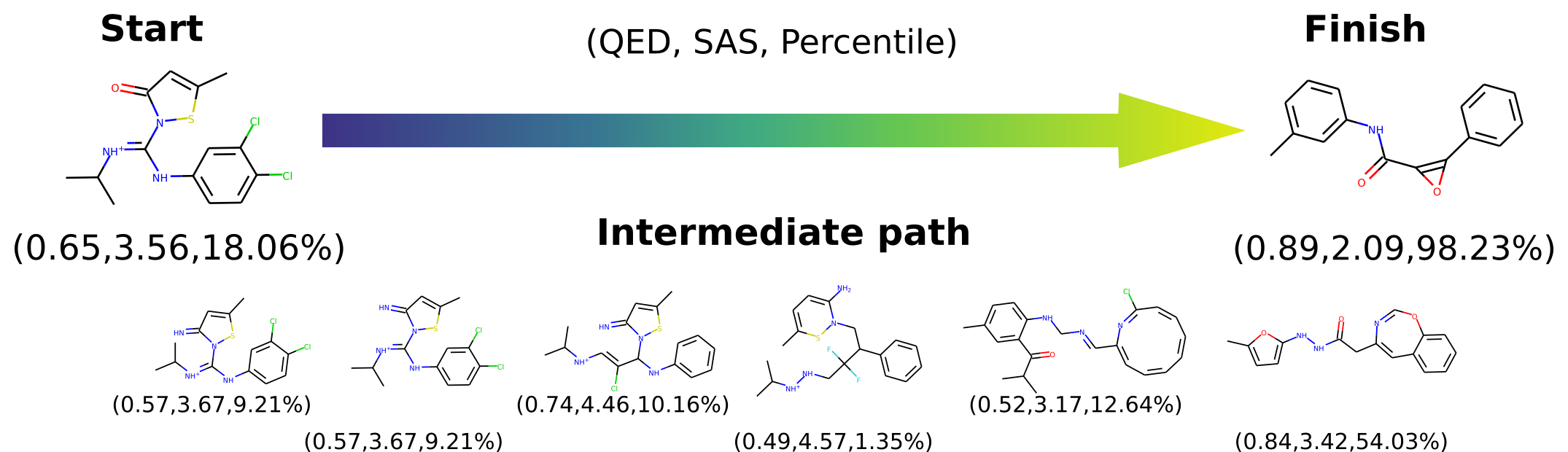}} 
	
	\caption{ Optimization results for the jointly trained autoencoder using $5 \times$ \texttt{QED} $-$ \texttt{SAS} as the objective function. Part (a) shows a box plot which compares the distribution of sampled molecules from normal random sampling, SMILES optimization via a common chemical transformation with a genetic algorithm, and from optimization on the trained gaussian process model with varying levels of accuracy/training points. To offset differences in computational cost between the random search and the optimization on the gaussian process model, the results of 400 iterations of random search were compared against the results of 200 iterations of optimization. This graph shows the combined results of four sets of trials. Part (b) shows the starting and ending points of several optimization runs on a PCA plot of latent space colored by the objective functon. Higlighted in black is the path illustrated in c). Part (c) shows a spherical interpolation between the actual start and finish molecules using a constant step size. The QED, SAS, and percentile score are reported for each molecule.}
    \label{fig:optimize_mols}
\end{figure}

From Figure \ref{fig:optimize_mols}a) we can see that the optimization with the Gaussian process model on the latent space representation consistently results in molecules with a higher percentile score than the two baseline search methods. Figure \ref{fig:optimize_mols}b) shows the path of one optimization from the starting molecule to the final molecule in the two-dimensional PCA representation, the final molecule ending up in the region of high objective value. Figure \ref{fig:optimize_mols}c)  shows molecules decoded along this optimization path using a Gaussian interpolation.

Performing this optimization on a Gaussian process (GP) model trained with 1,000 molecules leads to a slightly wider range of molecules as shown in Figure \ref{fig:optimize_mols}a). 
  Since the training set is smaller, the predictive power of the GP is lower which when optimizing in latent space, and as a result optimizes to several local minima instead of a global optimization.  In cases where it is difficult to define an objective that completely describes the desirable traits of the molecule, it may be better to use this localized optimization approach to reach a larger diversity of potential molecules.

\section{Conclusion}
We propose a new family of methods for exploring chemical space based on continuous encodings of molecules.
These methods eliminate the need to hand-build libraries of compounds and allow a new type of directed gradient-based search through chemical space. We observed high fidelity in reconstruction, the ability to capture characteristic features of a molecular training set into the generative model, good predictive power when training jointly an autoencoder and a predictor, and the ability to perform model-based optimization of molecules in the smoothed latent space.

There are several avenues for further improvement of this approach to molecular design.
In this work, we used a text-based molecular encoding, but using a graph-based autoencoder would have several advantages.
Forcing the decoder to produce valid SMILES strings makes the learning problem unnecessarily hard since the decoder must also implicitly learn which strings are valid SMILES.
An autoencoder that directly outputs molecular graphs is appealing since it could explicitly address issues of graph isomorphism and the problem of strings that do not correspond to valid molecular graphs. Building an encoder which takes in molecular graphs is straightforward through the use of off-the-shelf molecular fingerprinting methods, such as ECFP\cite{ECFP2010} or a continuously-parameterized variant of ECFP such as neural molecular fingerprints.\cite{duvenaud2015convolutional}
However, building a neural network which can output arbitrary graphs is an open problem. Further extensions of this work to use a explicitly defined grammar for SMILES instead of forcing the model to learn one\cite{kusner2017grammar} or to actively learn valid sequences\cite{Janz2017} are underway, as also is the application of adversarial networks for this task.\cite{guimaraes2017objective, Sanchez-Lengeling2017}

The autoencoder sometimes produced molecules that are formally valid as graphs but contain moieties that are not desirable because of stability or synthetic constraints.  Examples are acid chlorides, anhydrides, cyclopentadienes, aziridines, enamines, hemiaminals, enol ethers, cyclobutadiene, and cycloheptatriene. One option is to train the autoencoder with to predict properties related to steric constraints of other structural constraints. In general, the objective function to be optimized needs to capture as many desirable traits as possible and balance them to ensure that the optimizer focuses on genuinely desirable compounds.

The results reported in this work, and its application with carefully composed objective functions, have the potential to create new avenues for molecular design.

\section*{Methods}

\paragraph{Autoencoder architecture}
Strings of characters can be encoded into vectors using recurrent neural networks (RNNs). 
An encoder RNN can be paired with a decoder RNN to perform sequence-to-sequence learning.\cite{sutskever2014sequence}
We also experimented with convolutional networks for string encoding\cite{KalchbrennerACL2014} and observed improved performance.
This is explained by the presence of repetitive, translationally-invariant substrings that correspond to chemical substructures, e.g., cycles and functional groups.

Our SMILES-based text encoding used a subset of 35 different characters for ZINC and 22 different characters for QM9.
For ease of computation, we encoded strings up to a maximum length of 120 characters for ZINC and 34 characters for QM9, although in principle there is no hard limit to string length. Shorter strings were padded with spaces to this same length. We used only canonicalized SMILES for training to avoid dealing with equivalent SMILES representations.
The structure of the VAE deep network was as follows: For the autoencoder used for the ZINC dataset, the encoder used three 1D convolutional layers of filter sizes 9, 9, 10 and 9, 9, 11 convolution kernels, respectively, followed by one fully-connected layer of width 196. The decoder fed into three layers of gated recurrent unit (GRU) networks\cite{chung_2014_empirical} with hidden dimension of 488. % and 120 neurons?
For the model used for the QM9 dataset, the encoder used three 1D convolutional layers of filter sizes 2, 2, 1 and 5, 5, 4 convolution kernels, respectively, followed by one fully-connected layer of width 156. The three recurrent neural network layers each had a hidden dimension of 500 neurons.

The last layer of the RNN decoder defines a probability distribution over all possible characters at each position in the SMILES string.
This means that the writeout operation is stochastic, and the same point in latent space may decode into to different SMILES strings, depending on the random seed used to sample characters.
The output GRU layer had one additional input, corresponding to the character sampled from the softmax output of the previous time step and was trained using teacher forcing.\cite{williams_1989} This increased the accuracy of generated SMILES strings, which resulted in higher fractions of valid SMILES strings for latent points outside the training data, but also made training more difficult, since the decoder showed a tendency to ignore the (variational) encoding and rely solely on the input sequence. The variational loss was annealed according to sigmoid schedule after 29 epochs, running for a total 120 epochs.

For property prediction, two fully connected layers of 1000 neurons were used to predict properties from the latent representation, with a dropout rate of 0.2. For the algorithm trained on the ZINC dataset, the objective properties include logP, QED, SAS. For the algorithm trained on the QM9 dataset, the objective properties include HOMO energies, LUMO energies, and the electronic spatial extent (R$^2$). The property prediction loss was annealed in at the same time as the variational loss.
We used the Keras\cite{chollet_2015} and TensorFlow\cite{Tensorflow-2016} packages to build and train this model and the rdkit package for cheminformatics\cite{rdkit}.

\begin{acknowledgement}

This work was supported financially by the Samsung Advanced Institute of Technology.
The authors acknowledge the use of the Harvard FAS Odyssey Cluster and support from FAS Research Computing.
JNW acknowledges support from the National Science Foundation Graduate Research Fellowship Program under Grant No. DGE-1144152.
JMHL acknowledges support from the Rafael del Pino Foundation. RPA acknowledges support from the Alfred P.\ Sloan Foundation and NSF IIS-1421780. AAG acknowledges support from The Department of Energy, Office of Basic Energy Sciences under award DE-SC0015959.
We thank Dr. Anders Fr\o seth for his generous support of this work.

\end{acknowledgement}

\bibliography{biblio}

\newpage
\section*{Supplementary Materials}

 The code and full training data sets will be made available at \url{https://github.com/aspuru-guzik-group/chemical_vae}

\begin{table}[ht]
\centering
\begin{tabular}{|c|c|c|}
\hline 
Dataset & ZINC & QM \\ 
\hline 
Training set & 92.1 & 99.6 \\ 
\hline 
Test set & 90.7 & 99.4 \\ 
\hline 
ZINC & 91.0 & 1.4 \\ 
\hline 
eMolecules & 83.8 & 8.8 \\ 
\hline 
\end{tabular} 
\caption{Percentage of successfully decoding of latent representation after 1000 attempts for 1000 molecules from the traning set, 1000 validation molecules randomly chosen from ZINC and a 1000 validation molecules randomly chosen from eMolecules. Both VAEs perform very well for training data, and they are well transferable within molecules of the same class outside the training data, as evidence by the good validation performance of the ZINC VAE and the underperformance of the QM9 VAE against real-life small molecules.}
\label{tab:recovery statistics}
\end{table}

\begin{table}[ht]
\centering
\begin{tabular}{|c|c|c|}
\hline 
Dataset & ZINC & QM \\ 
\hline 
Decoding probability & 73.9 & 79.3 \\  
\hline 
\end{tabular} 
\caption{Percentage of 5000 randomly-selected latent points that decode to valid molecules after 1000 attempts}
\label{tab:random_sampling_statistics}
\end{table}

\newpage

\begin{figure}
\centering
\sidecaption{subfig:a}
\raisebox{-\height}{\includegraphics[width=0.45\columnwidth]{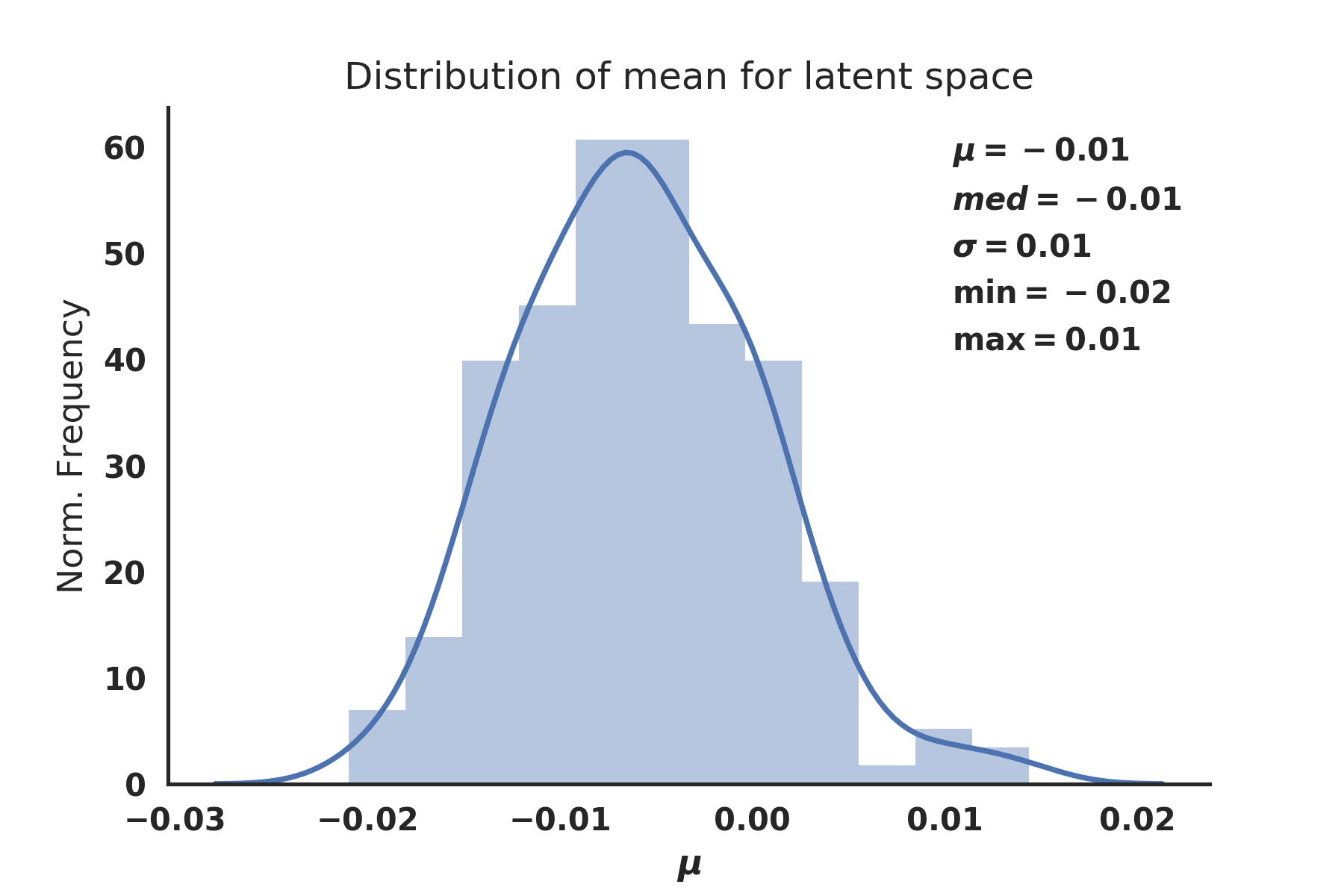}}
\sidecaption{subfig:b}
\raisebox{-\height}{\includegraphics[width=0.45\columnwidth]{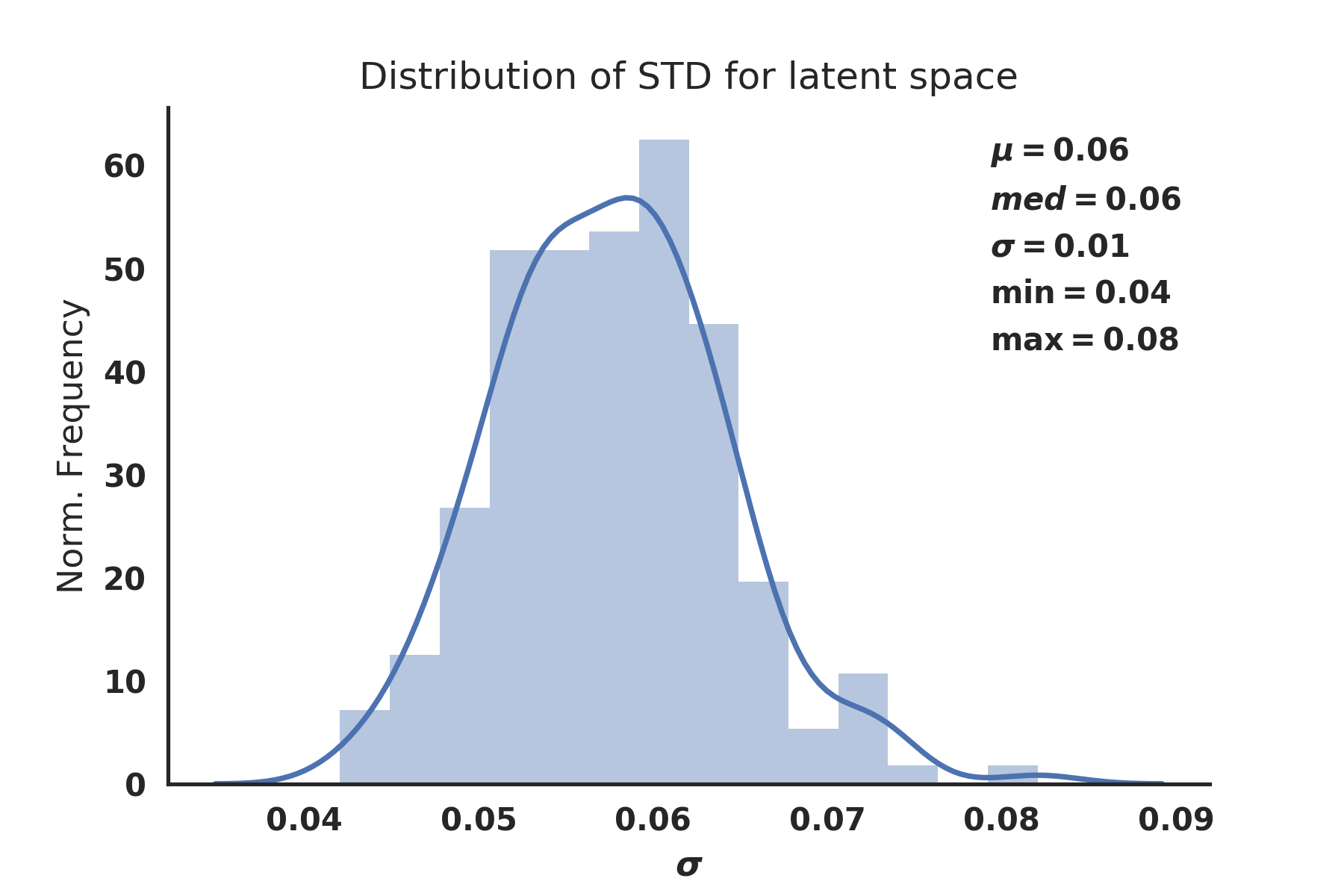}}

\sidecaption{subfig:c}
\raisebox{-\height}{\includegraphics[width=0.45\columnwidth]{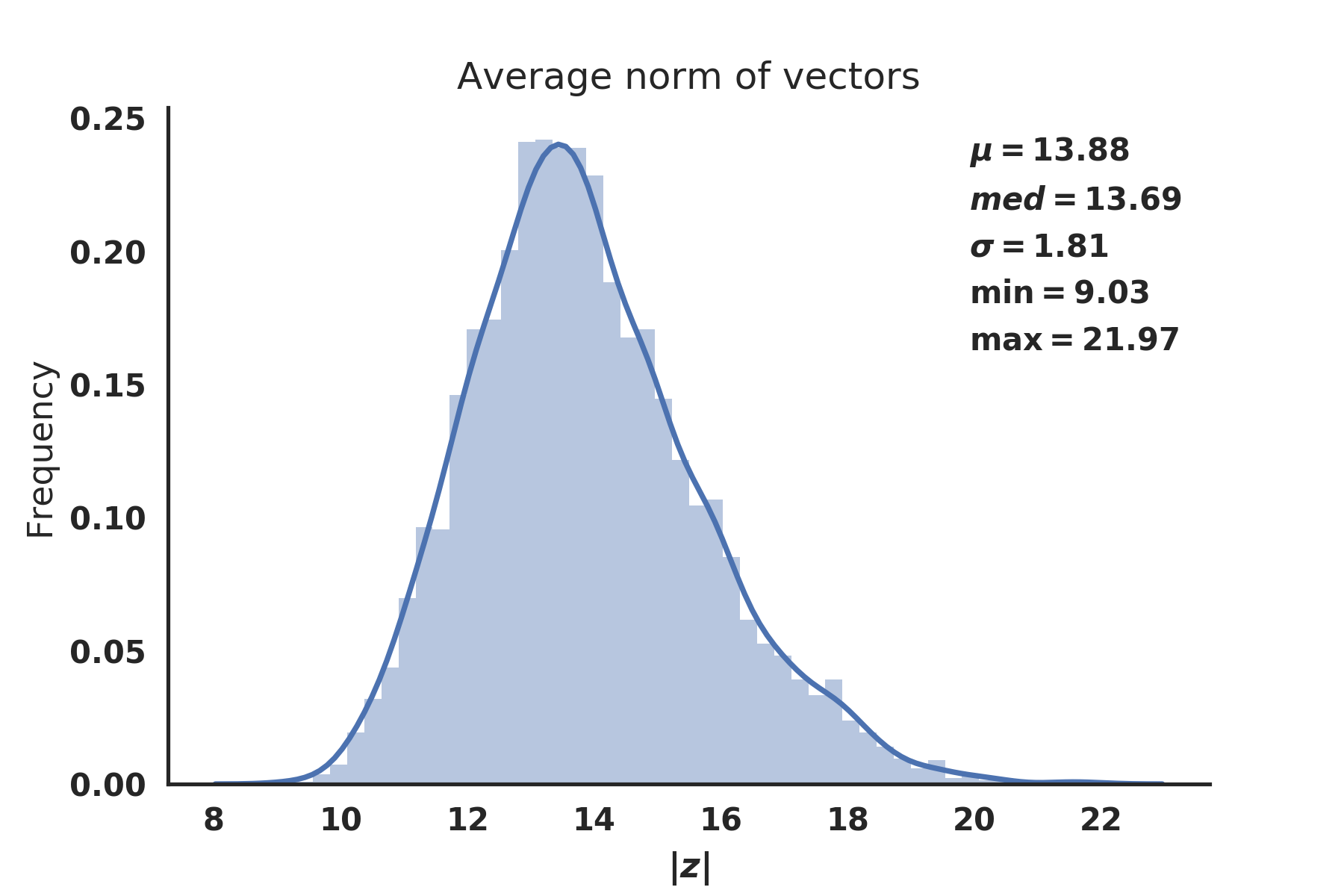}}
\sidecaption{subfig:d}
\raisebox{-\height}{\includegraphics[width=0.45\columnwidth]{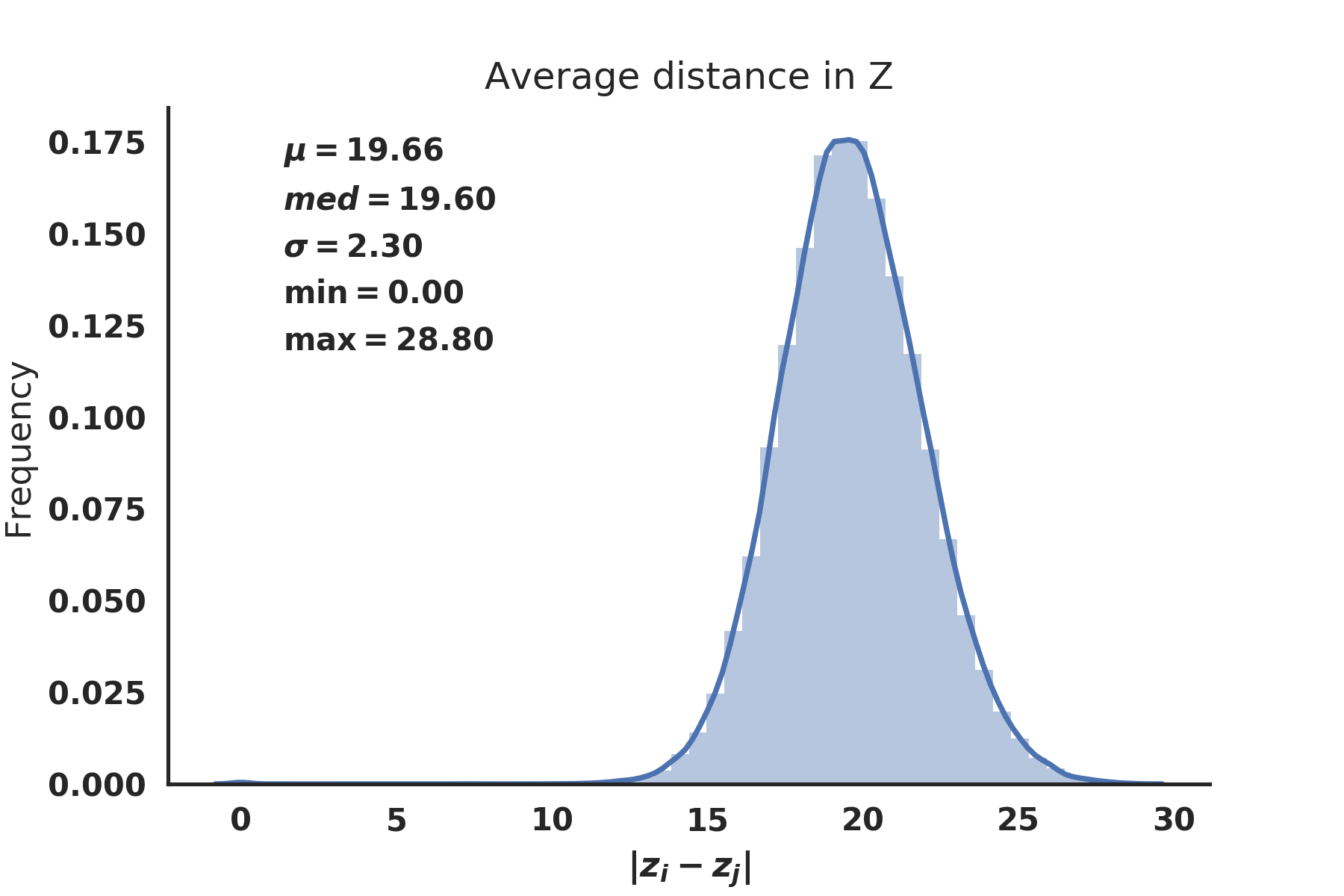}}
\caption{Distribution and statistics of (a) the mean of latent space coordinates (b) standard deviation of latent space coordinates (c) norm of latent space coordinates of the encoded representation of randomly selected molecules from the ZINC validation set. (d) Distribution of Euclidean distances between random pairs of validation molecules in the ZINC VAE }
\label{fig:ls_stats}
\end{figure}

\begin{figure}
\centering
\includegraphics[width=0.3\columnwidth]{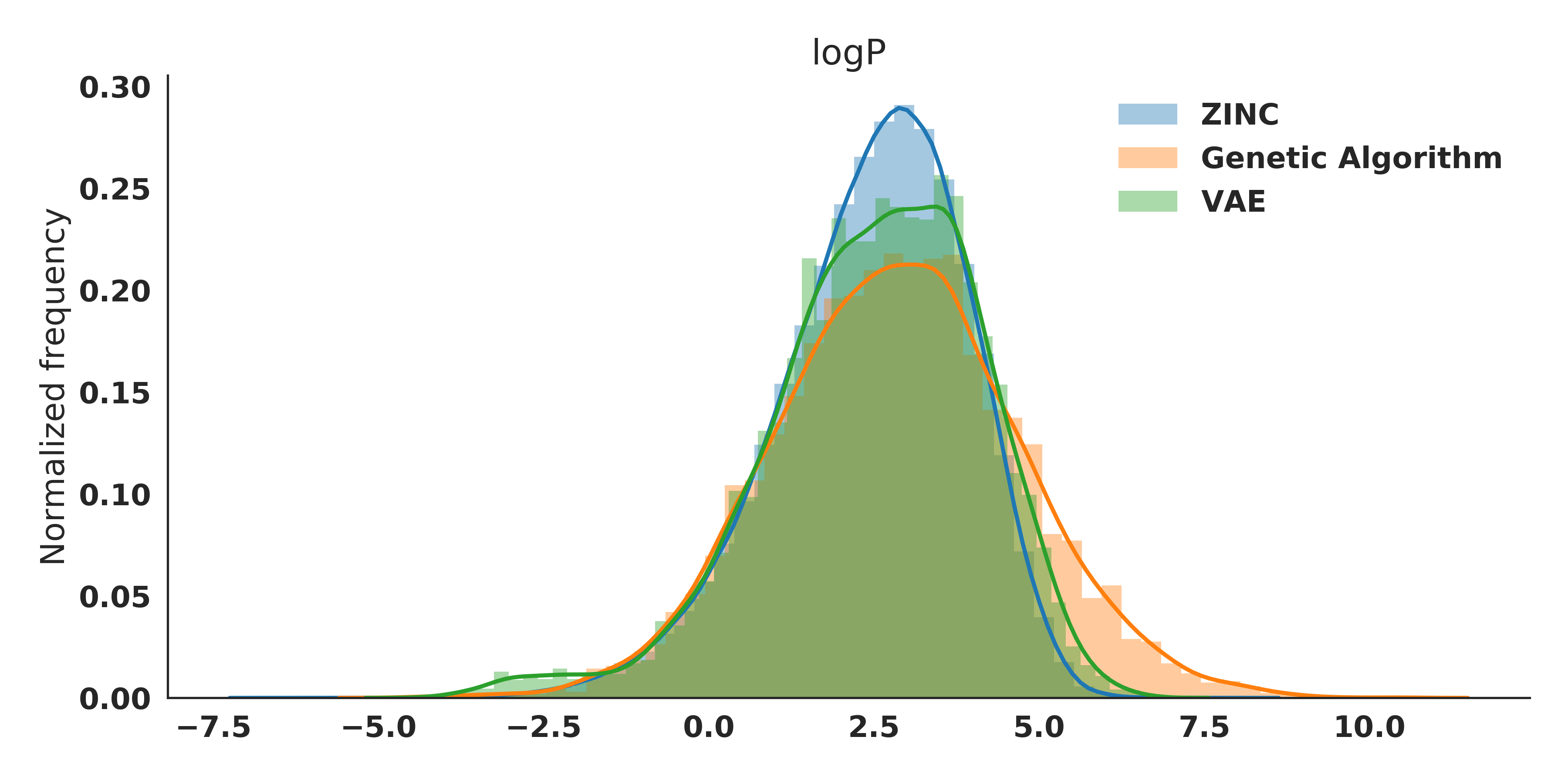}
\includegraphics[width=0.3\columnwidth]{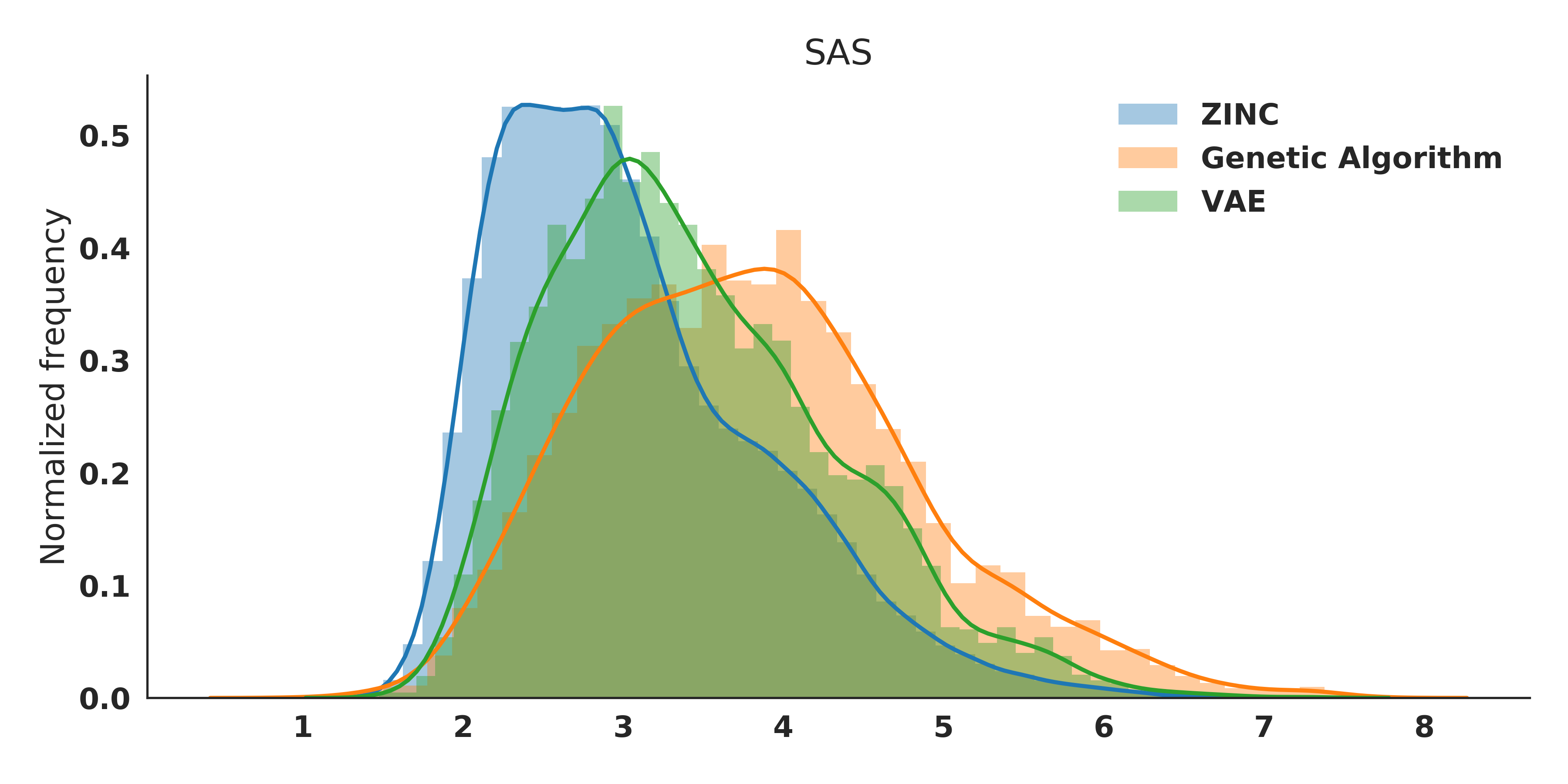}
\includegraphics[width=0.3\columnwidth]{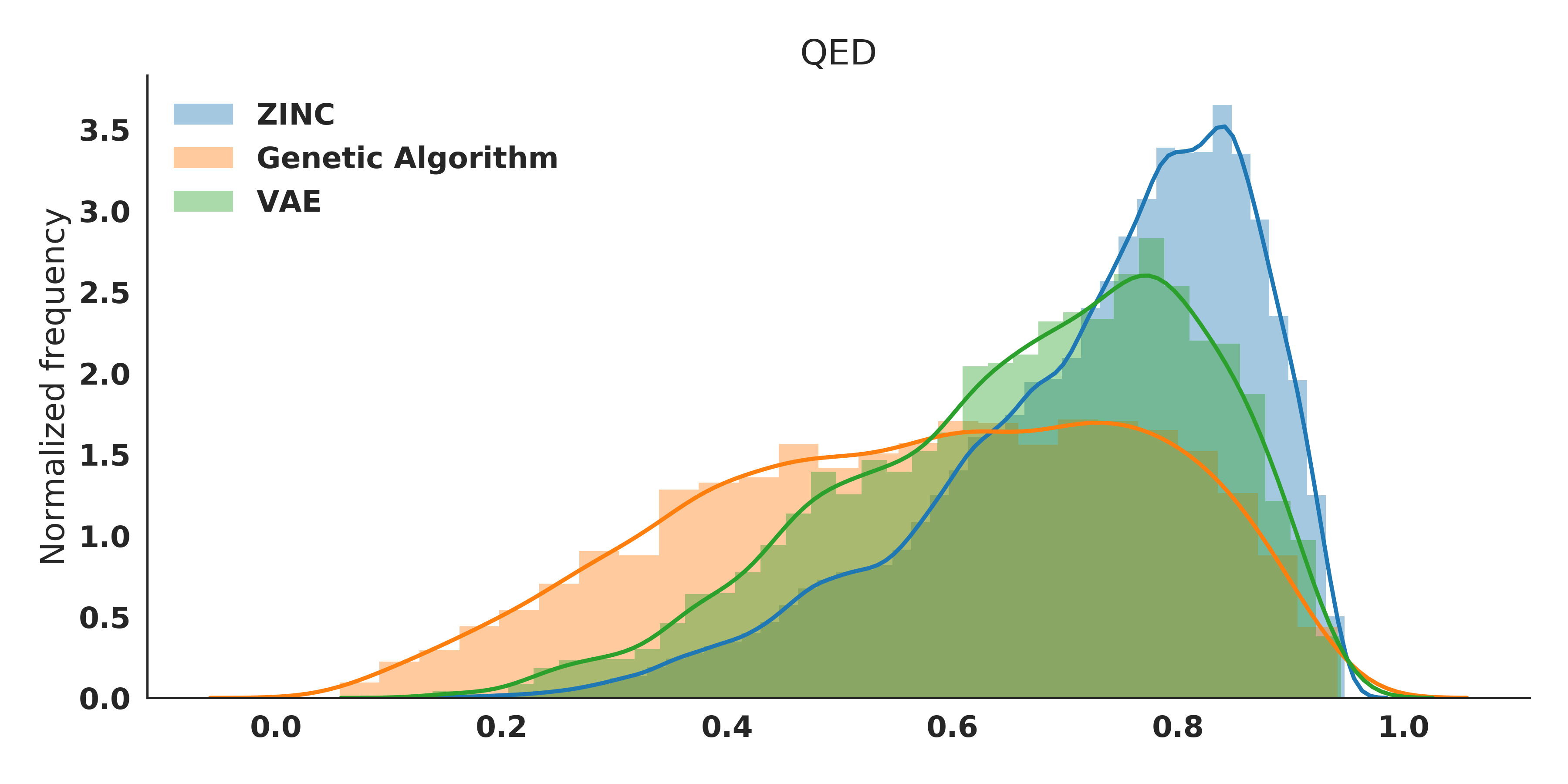}

\caption{Histograms and KDE plots of the distribution of properties utilized in the jointly trained autoencoder (LogP, SAS, QED). Used to further showcase results from Table 2. For each property we compare the distribution of the source data (ZINC), a generatic algorithm and the VAE.}
\label{fig:prop_dists}
\end{figure}

\begin{figure}
\centering
\includegraphics[width=\columnwidth]{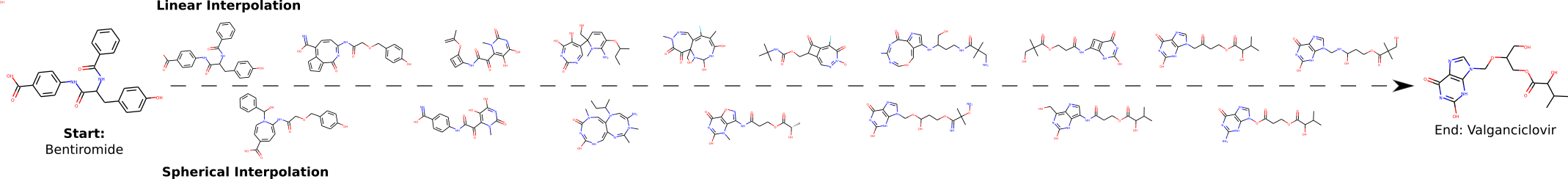}
\caption{Comparison of between linear and spherical interpolation paths between two randomly selected FDA approved drugs. A constant step size was used.}
\label{fig:interpol_1}
\end{figure}

\begin{figure}[h]
\centering
\includegraphics[width=0.9\columnwidth]{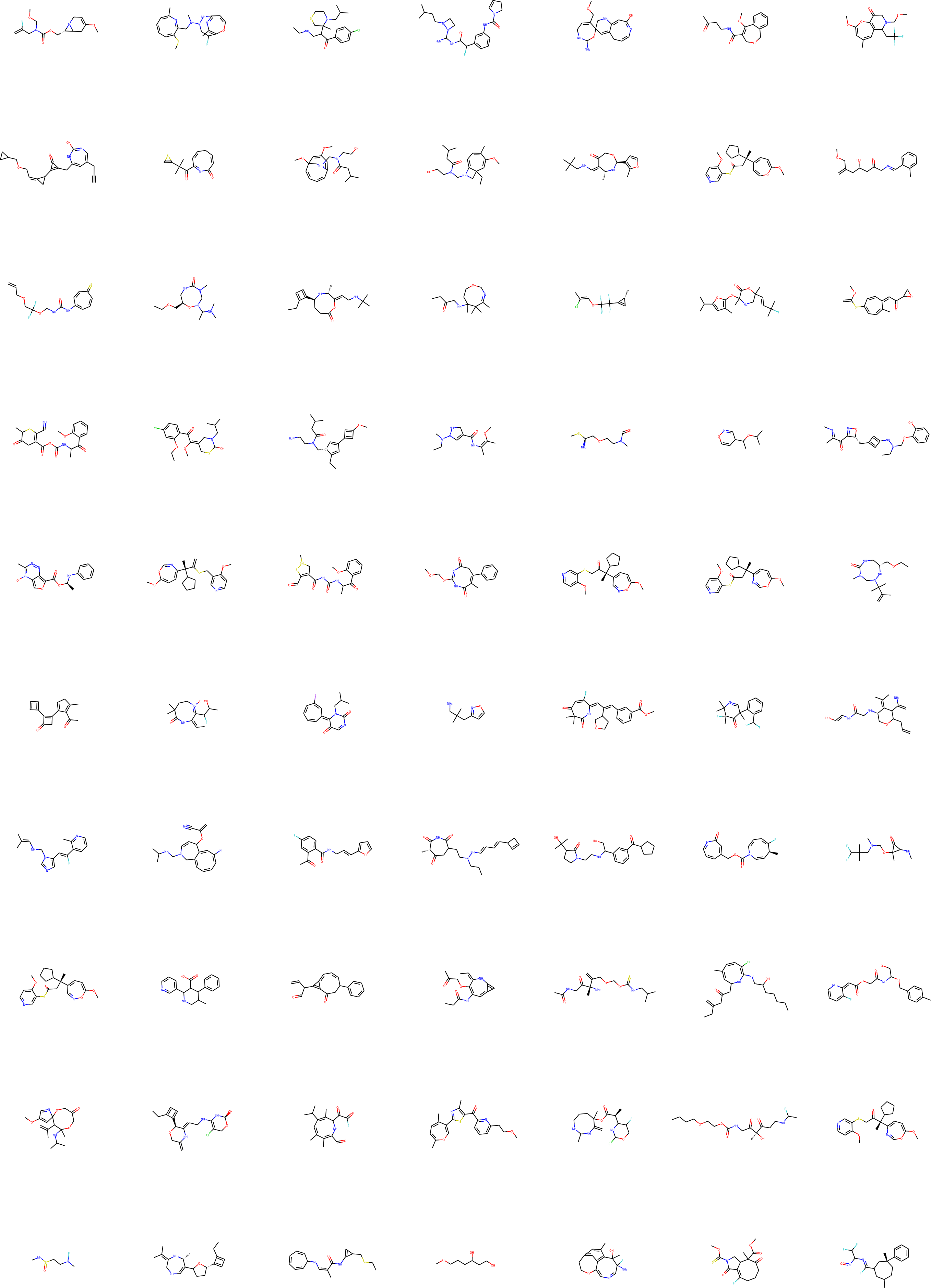} 
\caption{Molecules decoded from randomly-sampled points in the latent space of the ZINC VAE.}
\label{fig:random_3}
\end{figure}

\end{document}